\documentclass[conference]{IEEEtran}
\IEEEoverridecommandlockouts
\usepackage{cite}
\usepackage{amsmath,amssymb,amsfonts}
\usepackage{algorithmic}
\usepackage{graphicx}
\usepackage{textcomp}
\usepackage{xcolor}

\usepackage{array}
\usepackage{url}
\usepackage{multirow, tikz}
\usepackage{scalefnt}
\usepackage[capitalise, nameinlink]{cleveref}
\usepackage{flushend}

\newcommand{\R}{\mathbb{R}}

\def\BibTeX{{\rm B\kern-.05em{\sc i\kern-.025em b}\kern-.08em
		T\kern-.1667em\lower.7ex\hbox{E}\kern-.125emX}}

\begin{document}
	
	\title{Construction of a Surrogate Model: Multivariate Time Series Prediction with a Hybrid Model}
	
	\author{Clara Carlier\textsuperscript{(1)(2)},
		    Arnaud Franju\textsuperscript{(2)},
		    Matthieu Lerasle\textsuperscript{(1)},
		    Mathias Obrebski\textsuperscript{(2)}
		    \\
			\textsuperscript{(1)} Statistics Department of CREST, Palaiseau, 91120 France \\
			\textsuperscript{(2)} DEA-TDV team of the Renault group, Guyancourt, 78280 France
	}
	
	\maketitle
	
	\begin{abstract}
		Recent developments of advanced driver-assistance systems necessitate an increasing number of tests to validate new technologies. These tests cannot be carried out on track in a reasonable amount of time and automotive groups rely on simulators to perform most tests. The reliability of these simulators for constantly refined tasks is becoming an issue and, to increase the number of tests, the industry is now developing surrogate models, that should mimic the behavior of the simulator while being much faster to run on specific tasks. 
		
		In this paper we aim to construct a surrogate model to mimic and replace the simulator. We first test several classical methods such as random forests, ridge regression or convolutional neural networks. Then we build three hybrid models that use all these methods and combine them to obtain an efficient hybrid surrogate model.
	\end{abstract}
	
	
	\section{Introduction}
	
	% \begin{flushright} \textit{Context} \end{flushright}
	
	\IEEEPARstart{V}{alidation} and Certification of Autonomous Driver (AD) and Advanced Driver-Assistance Systems (ADAS) are highly sensitive applications that should be carried out very carefully. 
	The numerous on-board sensors in cars give access to a large amount of information. The regulations are strict and multiple. It is therefore necessary to carry out many on-track tests and to cover many kilometers. The Renault group has decided to develop digital platforms to model and simulate driving assistance and vehicle automation systems to create simulations. They will complete or even replace the real tests done on track in the validation process.
	
	In order to integrate the simulations into the validation and certification process, a digital twin to the physical autonomous vehicle must be created. Therefore, the simulator must be calibrated to generate data similar enough to the on-track tests. We want to develop a methodology that will gauge the quality of the simulator by comparing it to real on-track data and then calibrate and readjust it by changing the input parameters. Once recalibrated, the simulator should be able to generate time series closer to reality.
	
	\begin{flushright} \textit{Objectives and full process} \end{flushright}
	
\begin{figure*}[!t]
		\centering
		\includegraphics[trim=0cm 0cm 0cm 0cm, clip=true, width = 16cm]{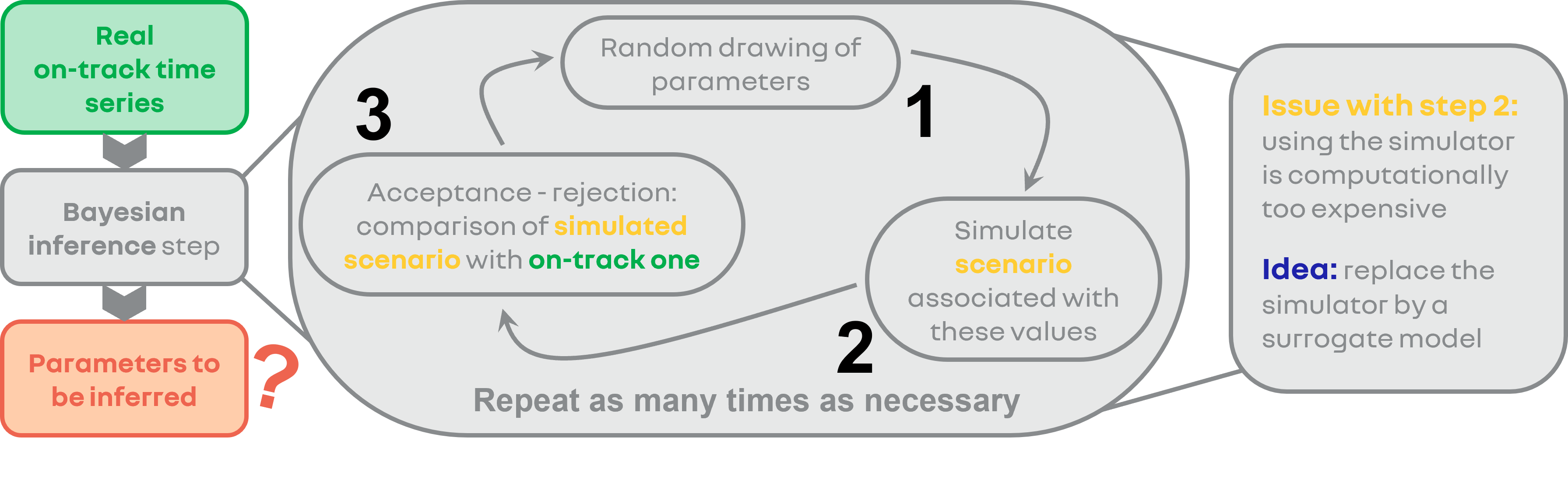}
		\caption{Summary of the general process. The three blocks on the left represent the inverse problem which consists in finding the values of the input parameters associated with the reference on-track test. The middle section describes how ABC methods work. The last part is about the issue we are facing and how we intend to solve it.}
		\label{fig:schema-demarche}
	\end{figure*}

	The proposed methodology to calibrate the simulator is articulated between the resolution of an inverse problem and a direct problem. % (Bayesian inference of parameters) (construction of a surrogate model)
	The whole process is similar to the one used in \cite{Loic-demarche-gen}, it is briefly described in \cref{fig:schema-demarche}. The main problem is the inverse one: we want to use Bayesian inference to recover the values of the input parameters associated to real time series realized on-track. We then get a posterior probability distribution for each parameter. 
	
	We focus more specifically on Approximate Bayesian Computation (ABC) methods which are likelihood-free inference schemes.
	In this approach, several steps are repeated iteratively and one of them requires to generate time series from a set of input parameters. 
	Concretely, we have a single reference on-track test describe by its real time series and its associated input parameters (like initial speed for example) called \textit{nominal values}. These nominal values are subject to uncertainties due to sensor noise or human tolerances for example. We will scan zones around these nominal parameters to determine which values would generate the most realistic simulation. 
	\\
	First, we (1) draw candidate parameters according to priors defined by these zones; then (2) we generate the time series associated to these candidate values, this is precisely what the simulator is for ; and finally, (3) we go to the acceptance-rejection step, if the generated series are close enough to the reference series, we accept the candidate values of the parameters, otherwise they are rejected. And (4) we repeat these three steps as many times as necessary or desired.
	
\begin{figure}[!t]
		\centering
		\includegraphics[trim=1.8cm 1.8cm 0.5cm 3cm, clip=true, width = 8.8cm]{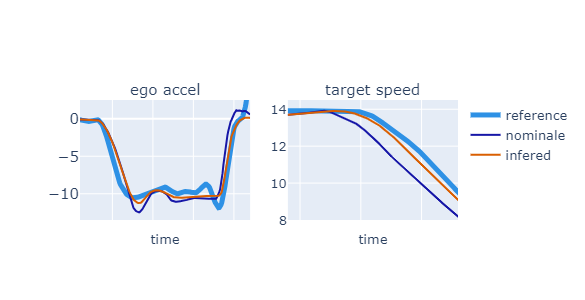}
		\caption{Reference on-track test and time series simulated with nominal parameters vs. inferred parameters}
		\label{fig:nominale-vs-inferred}
	\end{figure}

	\cref{fig:nominale-vs-inferred} shows the benefit of this parameter recalibration. We compare the real test to the simulations generated with nominal values vs. inferred parameters values. Predictions are closer to baseline values when using inferred rather than nominal parameters.
	%With the nominal values, we obtain a \textsc{rmse} of 1.47 while with the inferred and recalibrated parameters, the \textsc{rmse} is equal to 1.19.
	
	At this point, we are facing a problem. On step (2), the simulator is computationally too expensive to use in an iterative way. The idea is to develop a surrogate model that will mimic and replace the simulator and then perform this step during the inference. We now have to solve the direct problem by solving a classical learning problem involving multivariate time series.
	
	\begin{flushright} \textit{Surrogate model} \end{flushright}
	
	Surrogate models are largely used in all types of domains and contexts, they have already demonstrated their usefulness and efficiency using a wide variety of possible methods like: Polynomial Chaos Expansions \cite{pc-model}, Radial Basis Functions and Kriging \cite{rbf-model}, Bayesian Surrogate Models \cite{bayes-model}, Surrogate Response Surface Models \cite{resp-surf-model}, Artificial Neural Networks \cite{ann-model}.
	
	Surrogate models are also widely used in the automotive field and have demonstrated their accuracy in many applications like car seats \cite{surr-model-2}, suspension components \cite{surr-model-3}, human-product interaction \cite{surr-model-4} or autonomous vehicles validation \cite{rbf-model}.
	\\
	In this paper, we construct a surrogate model with supervised machine learning methods that mimics and replaces the Renault simulator by predicting the simulated time series. 
	
	The simulator is based on SCANeR$^{\textsc{tm}}$ studio software suite \cite{scaner}. SCANeR$^{\textsc{tm}}$ is dedicated to automotive and transport simulations. Among other things, it is designed to drive, test and validate AD/ADAS. It provides all the necessary tools to build a realistic virtual world by defining road environments, vehicle dynamics, traffic, weather, ..., etc. %sensors, real or virtual drivers, headlights, weather conditions and scenario scripting.
	SCANeR$^{\textsc{tm}}$ requires several input parameters to define the desired type of scenario and vehicle characteristics and then returns time series describing their behavior during the experiment. Each set of parameters generates a scenario described by several time series. 
	
	The surrogate model aims to predict all steps of the multiple output time series from a single set of parameters: this is not the classical context of time series forecasting where the goal is to predict the future of the time series from its past. The training dataset is a set of beforehand simulated time series output by SCANeR$^{\textsc{tm}}$ for various input parameters. The input parameters need to be carefully chosen so that the database correctly represents the desired parameter definition space.
	
\begin{figure}[!t]
		\centering
		\includegraphics[trim=1.8cm 1.8cm 0.5cm 3cm, clip=true, width = 8.8cm]{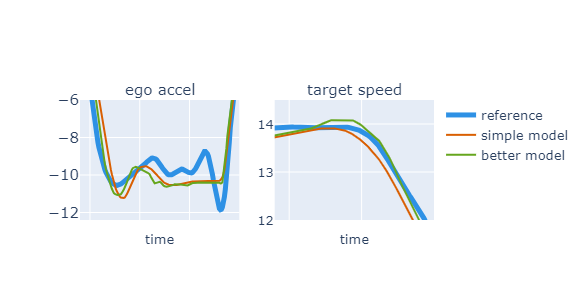}
		\caption{Reference on-track test and time series simulated from parameters inferred with a simple model vs. a better model}
		\label{fig:simple-vs-better}
	\end{figure}

	The surrogate model must be as accurate as possible: the simulations have to be close enough to the on-track tests to prove its quality and reliability, the surrogate model which replaces it must be as close as possible to the simulator's behavior to make its use viable in the general approach.
	\\
	Moreover, with a more accurate model, the final predictions are even better as can be appreciated in \cref{fig:simple-vs-better}.
	
	\bigskip
	\bigskip
	
	\begin{flushright} \textit{Mathematical description} \end{flushright}
	
	We consider a dataset $\big(\mathbf{x}_k, \mathbf{y}_k \big)$. Each $\mathbf{x}$ is a vector of $\R^{M}$ and corresponds to a set of parameters given to SCANeR$^{\textsc{tm}}$. Each $\mathbf{y}$ is a vector of $N$ time series $y_t^{(n)}$ of duration $T_n$ output by SCANeR$^{\textsc{tm}}$.
	
	We aim to predict the entire time series: given $\mathbf{x}$, we want to predict $\mathbf{y}$. These are linked through the deterministic simulator $\mathcal{S}^*$: for a given set of parameters, a single scenario is generated
	\begin{equation}
		\mathbf{y}_k = \mathcal{S}^* (\mathbf{x}_k)
	\end{equation}
	
	Constructing the surrogate model amounts to build a predictor $\widehat{\mathcal{S}}$ which returns all steps of time series. Given $\mathbf{x}$, we predict $\mathbf{y}$ by $\widehat{\mathcal{S}} (\mathbf{x})$. This corresponds to a classical supervised learning problem. 
	
	\section{Data description}
	
	In order to build the surrogate model, we construct a dataset with the SCANeR$^{\textsc{tm}}$ simulator. We look specifically for an emergency braking scenario involving two vehicles following each others and driving at  given initial constant speeds. The front vehicle (named \textbf{target}) brakes and the following one (named \textbf{ego}) activates its emergency braking to avoid collision.
	\\
	We give the simulator the values of the needed input parameters to define the type of scenario and the characteristics of the two vehicles, like initial speeds and braking efficiency of ego. The output time series will describe their behaviors: speed and acceleration of the ego vehicle, speed of the target vehicle and the distance between them. 
	
	% \newpage
	The dataset contained several scenarios and their time series generated by distinct sets of parameters. 
	Our goal is to use this dataset to build a surrogate model that is, a machine learning algorithm taking as input parameters and returning time series that should be close to those that SCANeR$^{\textsc{tm}}$ would have generated. 
	
	\subsection{Input parameters}
	
	For this specific scenario, SCANeR$^{\textsc{tm}}$ takes in input 7 parameters, divided into two groups. Scenario parameters define the initial speeds of the vehicles, the initial braking strength of the front vehicle and the initial distance between vehicles. Then, there is ego vehicle dynamic and environment parameters which define for example front and rear braking efficiency or Autonomous Emergency Braking (\textsc{aeb}) brake latency.
	
	\begin{table}[!t]
		\renewcommand{\arraystretch}{1.2}
		\caption{Intervals used to generate input parameters from uniform law}
		\label{tab:parameters-list}
		\centering
		\begin{tabular}{c l c c}
			\hline
			\textbf{Type} & \textbf{Parameter} & \textbf{Nominal} & \textbf{Interval}
			\\
			& & \textbf{value}
			\\
			\hline
			\textbf{scenario} & ego initial speed & $50$ & $[48, 52]$
			\\
			& target initial speed & $50$ & $[48, 52]$
			\\
			& target braking force & -6 & $[-8.5, -5]$
			\\
			& distance between the two & $40$ & $[38, 42]$
			\\
			\\
			\textbf{\textsc{Mada}} & front braking efficiency & $1$ & $[0.4, 1.6]$
			\\
			\textbf{\& \textsc{Env}} & rear braking efficiency & $1$ & $[0.4, 1.6]$
			\\
			& \textsc{aeb} brake latency & & $[0, 55]$
			\\
			\hline
		\end{tabular}
	\end{table}
	
	The simulator is deterministic so the variability in the training dataset only comes from the variability of input parameters. We draw each parameter from a uniform law, independently from each others. Intervals of each uniform law are defined by taking a percentage around the defined nominal values (\cref{tab:parameters-list}). These nominal values are given by the automotive certification authorities who define the values to be tested. 
	
\begin{figure}[!t]
		\centering
		\begin{tikzpicture}
			
			\node[inner sep = 0pt] (whitehead) at (0, 0)
			{\includegraphics[trim=2.5cm 10cm 2.5cm 2.8cm, clip=true, width = 7.5cm]{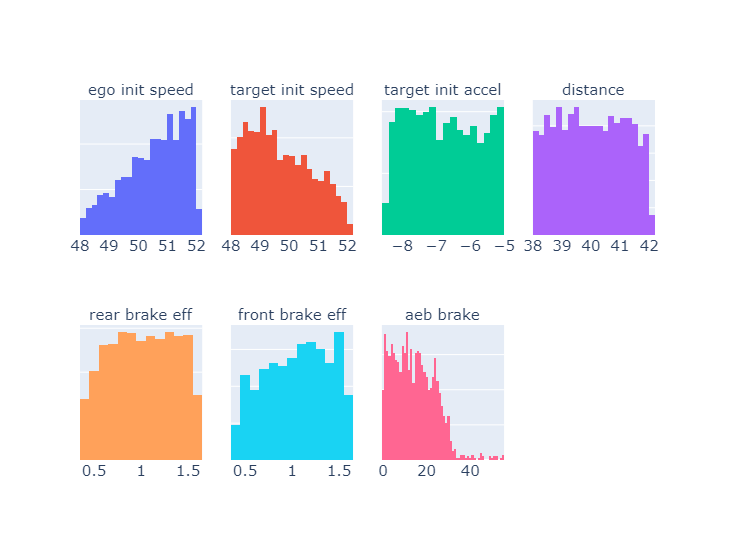}};
			\node[inner sep = 0pt] (whitehead) at (0.9, -2.4)
			{\includegraphics[trim=2.5cm 2.2cm 2.4cm 10.6cm, clip=true, width = 7.5cm]{figures/parameters_histogram.png}};
			
		\end{tikzpicture}
		\caption{Histograms of input parameters}
		\label{fig:parameters-histogram}
	\end{figure}

	After drawing each values, all sets of parameters are tested and some are automatically rejected. Indeed, there are some constraints on the parameter values and if they aren't complied, the simulator can't generate the associated scenario and crashes. For example, ego initial speed must be larger than target one. In \cref{fig:parameters-histogram}, we notice a huge loss of scenarios with too high target initial speeds or too low ego initial speeds.
	
	\subsection{Output time series}
	
\begin{figure}[!t]
		\centering
		\includegraphics[trim=2.2cm 1.8cm 2.8cm 2.8cm, clip=true,
		width = 7cm]{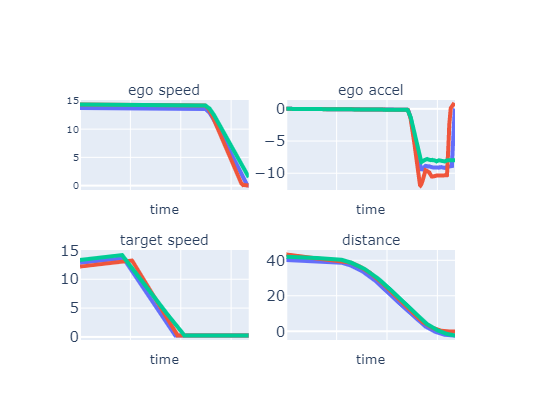}
		\caption{Time series corresponding to three experiments for three distinct sets of parameters}
		\label{fig:simu-example}
	\end{figure}

	\cref{fig:simu-example} shows the time series describing the speed and the acceleration of ego vehicle, the speed of target vehicle and the evolution of the distance between them. This gives an idea of the different profiles contained in our initial dataset.
	
	The dataset is divided into three parts: train (to fit the model), validation (to tune hyperparameters) and test (to provide an unbiased evaluation of a final model). Validation and test sets are composed of 100 scenarios each.
	
	\section{Surrogate model construction}
	
	In order to measure our risk, we consider the root mean squared error. For $u$ the vector containing the true values and $v$ the predicted values, the classical \textsc{rmse} is given by
	\begin{equation}
		\textsc{rmse} (u, v) = \sqrt{\dfrac{1}{n} \sum_{i=1}^{n} (u_i - v_i)^2}
	\end{equation}
	
	The \textsc{rmse} will be used to select the best method and the best approach.
	
	\subsection{Selection of the most promising method}
	
	\begin{table*}[!t]
		\renewcommand{\arraystretch}{1.3}
		\caption{\textsc{rmse} and computation times on train and validation sets for the seven methods \\ (*) \textit{laplacian kernel} ; (**) \textit{for 100 time series}}
		\label{tab:rmse-all-methods}
		\centering
		\begin{tabular}{|c|c|cccc|ccc|}
			\hline
			\multicolumn{2}{|c|}{} & $k$-\textsc{nn} & \textsc{krr} (*) & \textsc{cnn} & \textsc{df} & 1-\textsc{rf} & 4-\textsc{rf} & \textsc{pca-rf}
			\\
			\hline
			\textsc{rmse} & train & 9.23 & \textbf{0.04} & 1.22 & 4.26 & 1.24 & \textbf{0.71} & 1.95
			\\
			% \cline{2-9}
			$(\times 10^{-2})$ & validation & 30.15 & 7.21 & \textbf{2.31} & 7.05 & 7.27 & \textbf{3.69} & 12.35
			\\
			% \hline
			\hline
			\multicolumn{2}{|c|}{training time} & 0.05 sec & 0.22 sec & 59 min & 13 min & 42 sec & 53 sec & 7.42 sec
			\\
			% \hline
			\multicolumn{2}{|c|}{prediction time (**)} & 0.01 sec & 0.02 sec & 1.39 sec & 0.59 sec & 0.08 sec & 0.15 sec & 0.16 sec
			\\
			\hline
		\end{tabular}
	\end{table*}
	
	We realized a benchmark and tested different prediction methods to compare them, like $k$-nearest neighbors ($k$-\textsc{nn}), kernel ridge regression (\textsc{krr}, \textit{with laplacian kernel}) \cite{KRR}, simple convolutional neural networks (\textsc{cnn}), polynomial chaos expansion (\textsc{pce}) \cite{PCE-1, PCE-9}, random forests (\textsc{rf}), Deep Forest (\textsc{df}) \cite{DF}. 
	
	Concerning random forests, we developed a global model that predicts all the time series (1-\textsc{rf}) and another one obtained by training four distinct random forest models, one per time series (4-\textsc{rf}). We also tried to combine random forests with dimensionality reduction methods, like classical or functional PCA (\textsc{pca-rf}) \cite{fPCA-2, fPCA-1}.  
	
	\cref{tab:rmse-all-methods} summarizes the results obtained with each of these methods and compares the \textsc{rmse} and the computation times obtained. In terms of \textsc{rmse}, the $k$-\textsc{nn} algorithm produces the worst results. 
	The \textsc{pca-rf} results are not very good either, it reduces a bit the training time compared to \textsc{rf} but not the prediction time. 
	The results of \textsc{krr}, \textsc{df} and 1-\textsc{rf} are quite similar and a bit better. The \textsc{cnn} and the \textsc{4-rf} results are the best.
	
	\subsection{More detailed comparison}
	
	Let's now detail these results by calculating the \textsc{rmse} associated to each time series obtained with each method. The results are given in \cref{tab:rmse}.
	\\
	We now distinguish important differences that we could not notice with the previous table. Each method performs more or less well with each time series. In this case, \textsc{cnn} performs best for target speed and distance while \textsc{4-rf} performs best with the ego series.
	
	\begin{table}[!t]
		\renewcommand{\arraystretch}{1.3}
		\centering
		\caption{\textsc{rmse} for each time series on validation set \\ with the seven methods}
		\label{tab:rmse}
		\begin{tabular}{|c|cccc|c|c} 
			\cline{1-6}  
			\multirow{2}{*}{method} & ego & ego & target & \multirow{2}{*}{distance} & \multirow{2}{*}{mean} &
			\\
			& speed & accel & speed & & &
			\\
			\hline
			$k$-\textsc{nn} & 11.00 & 64.77 & 8.02 & 36.83 & 30.15 & \multicolumn{1}{c|}{}
			\\
			\textsc{krr} & 2.20 & 7.14 & 1.63 & 17.86 & 7.21 & \multicolumn{1}{c|}{$\star\star$}
			\\
			\textsc{cnn} & 0.18 & 3.85 & \textbf{1.06} & \textbf{4.16} & \textbf{2.31} & \multicolumn{1}{c|}{$\star\star\star$}
			\\
			\textsc{df} & 1.32 & 8.88 & 5.80 & 12.19 & 7.05 & \multicolumn{1}{c|}{$\star\star$}
			\\
			\hline
			1-\textsc{rf} & 1.36 & 9.54 & 4.86 & 13.31 & 7.27 & \multicolumn{1}{c|}{$\star\star$}
			\\
			4-\textsc{rf} & \textbf{0.12} & \textbf{1.47} & 2.34 & 10.84 & 3.69 & \multicolumn{1}{c|}{$\star\star\star$}
			\\
			\textsc{pca-rf} & 2.33 & 20.53 & 5.58 & 20.96 & 12.35 & \multicolumn{1}{c|}{$\star$}
			\\
			\hline
		\end{tabular}
	\end{table}
	
	\begin{figure}[!t]
		\centering
		\includegraphics[trim=1.8cm 1.8cm 0.5cm 3cm, clip=true,
		width = 8cm]{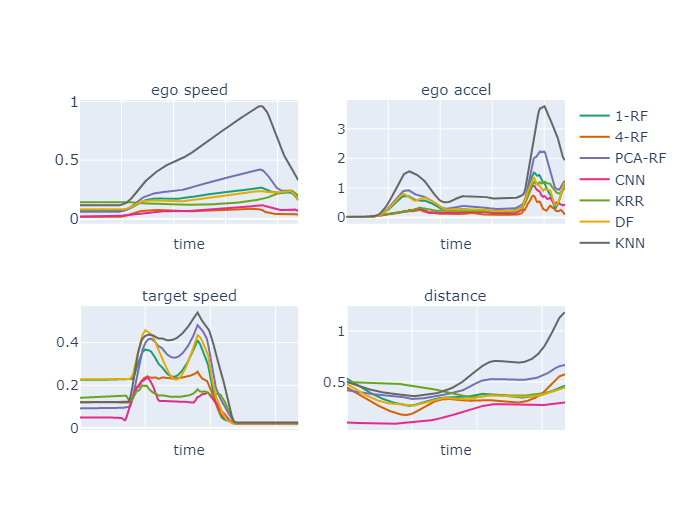}
		\caption{Mean \textsc{rmse}: for each method, we compute the mean \textsc{rmse} obtained at each time step on validation set}
		\label{fig:rmse-each-time}
	\end{figure}
	
	We then want to detail further the error values: for all methods, the average \textsc{rmse} is calculated for each time step. \cref{fig:rmse-each-time} represents this mean \textsc{rmse}. 
	\\
	We can now distinguish which method performs better at each time step. For distance, \textsc{cnn} clearly outperforms other methods, but it's not so clear for the other series. These results suggest to build hybrid models. 
	
	\section{Hybrid and aggregated models}
	
	We build three new models: two with hybrid approach and one with aggregation. The first two consist in choosing the best method at each time step and the third one performs a mixture of methods by giving them different weights computed with an expert aggregation.
	
	The description of the three models is:
	\begin{itemize}
		\item Hybrid 1: for each time step, we select the best method among the seven proposed;
		\item Hybrid 2: we keep the three most used methods in Hybrid 1 (\textsc{cnn}, 4-\textsc{rf} and \textsc{pca-rf}) and select the best one for each time step;
		\item Aggregated: it is built with an Exponential Weighted Aggregation (\textsc{ewa}) \cite{m2-jaouad}. 
	\end{itemize}
	
\begin{figure*}[!t]
		\centering
		{\scalefont{0.7}
			\begin{tikzpicture}
				
				\node[inner sep = 0pt] (whitehead) at (0, 2)
				{\includegraphics[trim=1.5cm 2.5cm 3cm 2.7cm, clip=true, height=2.4cm, width = 17cm]{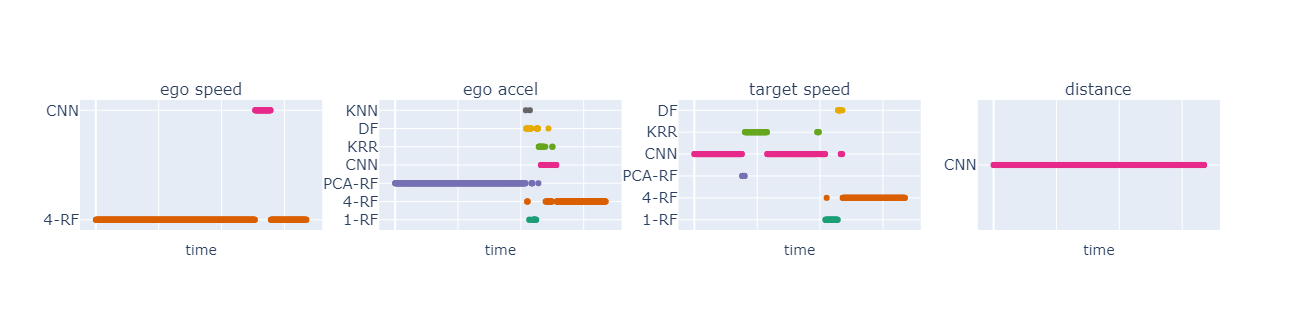}};
				\node[inner sep = 0pt] (whitehead) at (0, -1.1)
				{\includegraphics[trim=1.5cm 1.5cm 3cm 3.5cm, clip=true, height=2.4cm, width = 17cm]{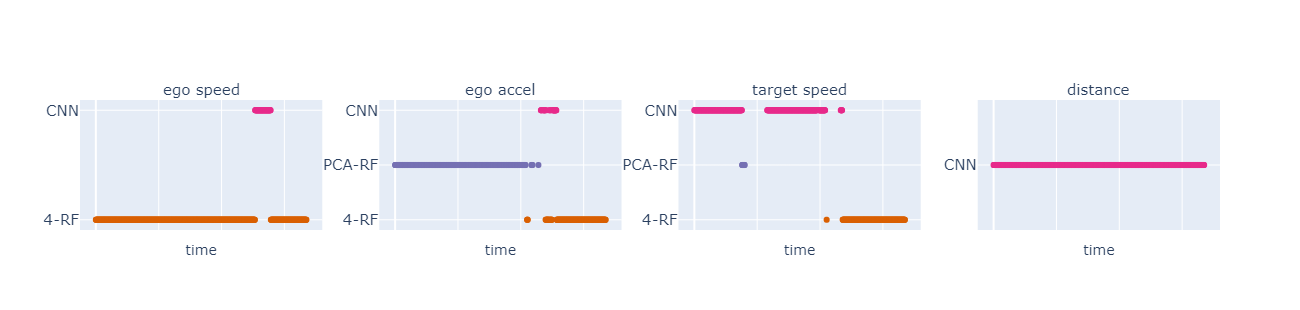}};
				
				\node[inner sep = 0pt] (whitehead) at (0.52, 0.55)
				{\includegraphics[trim=2.7cm 2.8cm 1.5cm 3.4cm, clip=true, width = 17.05cm]{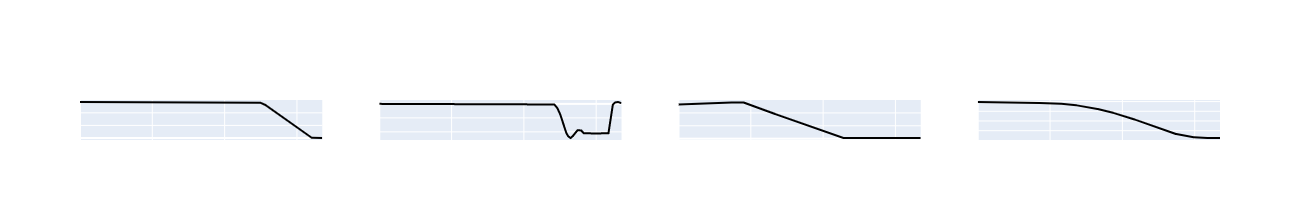}};
				
				\node at (9.3, 1.8) 
				{Hybrid 1 (*)}; % [rotate = -90]
				\node at (9.3, -0.6) 
				{Hybrid 2 (**)};
				
		\end{tikzpicture}}
		\caption{Selected method at each time step on validation set. (*) Hybrid 1: among the 7 methods ; (**) Hybrid 2: among the 3 best methods}
		\label{fig:selected-methods}
	\end{figure*}

\begin{figure*}[!t]
		\centering
		\includegraphics[trim=1.5cm 1.5cm 0.5cm 2.7cm, clip=true,
		width = 18.8cm]{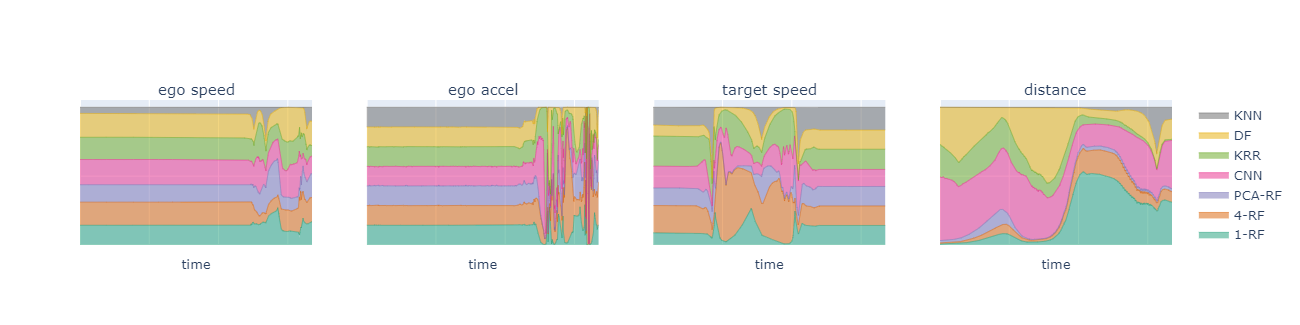}
		\caption{Associated weights for each expert at each time step on validation set}
		\label{fig:poids-experts-each-time}
	\end{figure*}

	\cref{fig:selected-methods} shows which method performs best at each time step and the weights given to each method for each time step are shown in \cref{fig:poids-experts-each-time}.
	
	\subsection{Numerical results}
	
	We now summarize all numerical results obtained with these three approaches. First on the validation set which is used to calibrate: the choice of the methods and the weights for each time step. Then, on the test set to confirm the results.
	\medskip
	
	\subsubsection{Validation (\cref{tab:rmse-validation})}
	
	The three new models are better than the \textsc{cnn} and \textsc{4-rf} models. The aggregated one gets better overall results. Hybrid 1 and Hybrid 2 are quite similar. Let's now see how these results generalize to the test set.
	
	\medskip
	
	\subsubsection{Test (\cref{tab:rmse-test})}
	
	\begin{table}[!t]
		\renewcommand{\arraystretch}{1.3}
		\centering
		\caption{\textsc{rmse} for each time series on validation set \\ with cnn, 4-Rf and three hybrid models}
		\label{tab:rmse-validation}
		\begin{tabular}{|c|cccc|c|} 
			\hline   
			method & ego speed & ego accel & target speed & distance & mean
			\\
			\hline
			\textsc{cnn} & 0.18 & 3.85 & {1.06} & {4.16} & {2.31} %& \multicolumn{1}{c|}{$\star \star \star$}
			\\
			4-\textsc{rf} & 0.12 & {1.47} & 2.34 & 10.84 & 3.69 %& \multicolumn{1}{c|}{$\star \star \star$}
			\\
			\hline
			Hybrid 1 & 0.11 & 1.46 & 0.93 & 4.16 & 1.66 %& \multicolumn{1}{c|}{$\star \star \star$}
			\\
			Hybrid 2 & 0.11 & 1.46 & 1.04 & 4.16 & 1.69 %& \multicolumn{1}{c|}{$\star \star \star$}
			\\
			Aggregated & \textbf{0.07} & \textbf{0.59} & \textbf{0.24} & \textbf{1.36} & \textbf{0.56} %& \multicolumn{1}{c|}{$\star \star \star$}
			\\
			\hline
		\end{tabular}
	\end{table}
	
	\begin{table}[!t]
		\renewcommand{\arraystretch}{1.3}
		\centering
		\caption{\textsc{rmse} for each time series on test set \\ with cnn, 4-rf and three hybrid models}
		\label{tab:rmse-test}
		\begin{tabular}{|c|cccc|c|} 
			\hline        
			method & ego speed & ego accel & target speed & distance & mean
			\\
			\hline
			\textsc{cnn} & 0.23 & 3.34 & {1.00} & {2.52} & {1.77} %& \multicolumn{1}{c|}{$\star \star \star$}
			\\
			4-\textsc{rf} & 0.13 & {1.36} & 2.36 & 9.84 & 3.42 %& \multicolumn{1}{c|}{$\star \star \star$}
			\\
			\hline
			Hybrid 1 & 0.12 & 1.35 & 1.12 & 2.52 & 1.28 %& \multicolumn{1}{c|}{$\star \star \star$}
			\\
			Hybrid 2 & 0.12 & 1.35 & \textbf{1.00} & 2.52 & \textbf{1.25} %& \multicolumn{1}{c|}{$\star \star \star$}
			\\
			Aggregated & 0.50 & 3.66 & 1.08 & 3.38 & 2.16 %& \multicolumn{1}{c|}{$\star \star \star$}
			\\
			\hline
		\end{tabular}
	\end{table}
	
	On the test set, the results of the three hybrid models are still better than the \textsc{cnn} and \textsc{4-rf} ones. 
	\\
	The aggregated model is no longer the best, it is even worse than the \textsc{cnn} model. The weight distribution cannot be generalized and is too specific to the validation set. 
	
	With the Hybrid 1 model, the prediction of target speed is deteriorated. From these results, we can see the advantage of the Hybrid 2 model: for each time series, it improves predictions although the choice of methods was made on the validation set. Whatever the database used to calibrate, it is the Hybrid 2 model that generalizes best.
	\medskip
	
	\subsubsection{Computation times (\cref{tab:hybrid-agg-times})}
	
	\begin{table}[!t]
		\renewcommand{\arraystretch}{1.3}
		\centering
		\caption{Computation times for cnn, 4-RF and three hybrid models \\ (*) \textit{for 100 time series} ; $(+)$ : \textit{add the prediction times of each method} }
		\label{tab:hybrid-agg-times}
		\begin{tabular}{|c|cc|} 
			\hline    
			time & \textsc{cnn} & \textsc{4-rf}
			\\
			\hline
			training & 59 min & 53 sec 
			\\
			prediction (*) & 1.39 sec & 0.15 sec
			\\
			\hline
		\end{tabular}
		
		\medskip
		
		\begin{tabular}{|c|ccc|} 
			\hline    
			time & Hybrid 1 & Hybrid 2 & Aggregated
			\\
			\hline
			weight computation $(+)$ & 0.29 sec & 0.18 sec & 2 min 13 sec
			\\
			prediction (*) & 10.25 sec & 8.59 sec & 2 min 17 sec (+)
			\\
			\hline
		\end{tabular}
	\end{table}
	
	\noindent The aggregated model is very time consuming, although it is still faster than SCANeR$^{\textsc{tm}}$. This model does not bring anything more than the other approaches either in terms of accuracy or performance. The two hybrid models improve the results but still multiply the prediction times by 10. Depending on the case, it will be necessary to decide which approach is the most judicious.
	
	\section{Conclusion}
	
	The objective was to replace the simulator which is computationally too expansive: an iteration on the simulator SCANeR$^{\textsc{tm}}$ used at Renault takes at least 15 minutes, which becomes prohibitive when repeated simulations are required, as it will be the case in the ABC algorithm we intend to develop. 
	
	After testing several methods and different approaches, kernel ridge regression, convolutional neural network and random forests stood out. 
	The computation times of these surrogate models are much more reasonable. 
	
	A closer look at the prediction errors shows that some methods are more efficient in predicting certain time series and that different can even be more interesting at different time step of each time series.
	
	We thus built three new models taking advantage of these preliminary remarks:
	\begin{itemize}
		\item Hybrid 1 selects the method with the lowest \textsc{rmse} at each time step;
		\item Hybrid 2 selects the method with the lowest \textsc{rmse} at each time step only among the three globally more efficient methods in Hybrid 1, which seems to avoid overfitting.
		\item Aggregated model uses an aggregation of the experts with an exponential weighted algorithm.
	\end{itemize}
	
	These three hybrid models provide a clear improvement in predictions over the basic methods. The aggregated model seems to generalize worse on the test set. Hybrid 1 is quite good on both validation and test set but Hybrid 2 generalizes better. 
	
	Concerning training times, these three new models are clearly more time consuming. But the \textsc{cnn} model trains at least during one hour. So adding few minutes will not be restrictive. 
	\\
	However, the prediction time is deteriorated, which might become restrictive for their use in ABC methods. The computation times in \cref{tab:total-times} shows that there will be a tradeoff to find for practitioners between accuracy and computation time.
	
	\begin{table}[!t]
		\centering
		\renewcommand{\arraystretch}{1.3}
		\caption{Computation time to generate 50.000 simulations one by one}
		\label{tab:total-times}
		\begin{tabular}{ccc} 
			\hline  
			\textbf{\textsc{4-rf}} & \textbf{Hybrid 2} & \textbf{SCANeR$^{\textsc{tm}}$}
			\\
			\hline
			1 minute & 1 hour & 5 days
			\\
			\hline
		\end{tabular}
	\end{table}
	
	To conclude, we easily built an overall reasonable model using random forests.
	To improve this benchmark, we specify which method to use at each time step of each time series. We emphasized the cost of this refinement and leave the final choice to the user's time constraints.
	
	\bibliographystyle{IEEEtran}
	\bibliography{IEEEabrv,biblio-file}

\begin{thebibliography}{10}
\providecommand{\url}[1]{#1}
\csname url@samestyle\endcsname
\providecommand{\newblock}{\relax}
\providecommand{\bibinfo}[2]{#2}
\providecommand{\BIBentrySTDinterwordspacing}{\spaceskip=0pt\relax}
\providecommand{\BIBentryALTinterwordstretchfactor}{4}
\providecommand{\BIBentryALTinterwordspacing}{\spaceskip=\fontdimen2\font plus
\BIBentryALTinterwordstretchfactor\fontdimen3\font minus
  \fontdimen4\font\relax}
\providecommand{\BIBforeignlanguage}[2]{{%
\expandafter\ifx\csname l@#1\endcsname\relax
\typeout{** WARNING: IEEEtran.bst: No hyphenation pattern has been}%
\typeout{** loaded for the language `#1'. Using the pattern for}%
\typeout{** the default language instead.}%
\else
\language=\csname l@#1\endcsname
\fi
#2}}
\providecommand{\BIBdecl}{\relax}
\BIBdecl

\bibitem{Loic-demarche-gen}
L.~Giraldi, O.~P.~L. Maître, K.~T. Mandli, C.~N. Dawson, I.~Hoteit, and O.~M.
  Knio, ``Bayesian inference of earthquake parameters from buoy data using a
  polynomial chaos-based surrogate,'' \emph{Comput Geosci 21}, pp. 683--699,
  2017.

\bibitem{pc-model}
I.~Sraj, K.~Mandli, O.~Knio, C.~Dawson, and I.~Hoteit, ``{Quantifying
  uncertainties in Fault Slip Distribution During the T\=ohoku Tsunami using
  Polynomial Chaos},'' \emph{Ocean Dynamics}, vol.~67, 2016.

\bibitem{rbf-model}
H.~Beglerovic, M.~Stolz, and M.~Horn, ``Testing of autonomous vehicles using
  surrogate models and stochastic optimization,'' \emph{2017 IEEE 20th
  International Conference on Intelligent Transportation Systems (ITSC)}, pp.
  1--6, 2017.

\bibitem{bayes-model}
E.~B. Ford, A.~V. Moorhead, and D.~Veras, ``A bayesian surrogate model for
  rapid time series analysis and application to exoplanet observations,''
  \emph{Bayesian Analysis}, vol.~6, no.~3, pp. 475--499, 2011.

\bibitem{resp-surf-model}
S.~A. Mattis and B.~Wohlmuth, ``Goal-oriented adaptive surrogate construction
  for stochastic inversion,'' \emph{Computer Methods in Applied Mechanics and
  Engineering}, vol. 339, pp. 36--60, 2018.

\bibitem{ann-model}
Z.~Xu, C.~Yu, H.~Sun, and Z.~Yang, ``The response of sediment phosphorus
  retention and release to reservoir operations: Numerical simulation and
  surrogate model development,'' \emph{Journal of Cleaner Production}, vol.
  271, p. 122688, 2020.

\bibitem{surr-model-2}
J.~Long, Y.~Liao, and P.~Yu, ``Multi-response weighted adaptive sampling
  approach based on hybrid surrogate model,'' \emph{IEEE Access}, vol.~9, pp.
  45\,441--45\,453, 2021.

\bibitem{surr-model-3}
R.~Jiang, Z.~Jin, D.~Liu, and D.~Wang, ``{Multi-Objective Lightweight
  Optimization of Parameterized Suspension Components Based on NSGA-II
  Algorithm Coupling with Surrogate Model},'' \emph{Machines}, vol.~9, no.~6,
  2021.

\bibitem{surr-model-4}
S.~Ahmed, M.~S. Gawand, L.~Irshad, and H.~O. Demirel, ``{Exploring the Design
  Space Using a Surrogate Model Approach With Digital Human Modeling
  Simulations},'' \emph{International Design Engineering Technical Conferences
  and Computers and Information in Engineering Conference}, vol.~1B, 2018.

\bibitem{scaner}
{AV Simulation, SCANeR$^{\textsc{tm}}$ studio},
  \url{https://www.avsimulation.com/scaner-studio/}.

\bibitem{KRR}
P.~Exterkate, P.~J. Groenen, C.~Heij, and D.~van Dijk, ``Nonlinear forecasting
  with many predictors using kernel ridge regression,'' \emph{International
  Journal of Forecasting}, vol.~32, no.~3, pp. 736--753, 2016.

\bibitem{PCE-1}
T.~Crestaux, O.~{Le Maître}, and J.-M. Martinez, ``Polynomial chaos expansion
  for sensitivity analysis,'' \emph{Reliability Engineering \& System Safety},
  vol.~94, no.~7, pp. 1161--1172, 2009, special Issue on Sensitivity Analysis.

\bibitem{PCE-9}
G.~Blatman and B.~Sudret, ``Adaptive sparse polynomial chaos expansion based on
  least angle regression,'' \emph{Journal of Computational Physics}, vol. 230,
  no.~6, pp. 2345--2367, 2011.

\bibitem{DF}
Z.-H. Zhou and J.~Feng, ``Deep forest,'' \emph{National Science Review},
  vol.~6, no.~1, pp. 74--86, 2018.

\bibitem{fPCA-2}
J.~Wohlenberg, ``{Functional Principal Component Analysis and Functional
  Data},'' \emph{Toward Datascience}, June 2021.

\bibitem{fPCA-1}
H.~{Shang}, ``A survey of functional principal component analysis,'' \emph{AStA
  Advances in Statistical Analysis}, pp. 121--142, 2014.

\bibitem{m2-jaouad}
J.~Mourtada, ``{Prédiction séquentielle par agrégation d’experts},''
  \emph{Master dissertation}, 2016.

\end{thebibliography}
	
\end{document}